\documentclass[journal]{IEEEtran}
\usepackage{graphicx}
\usepackage{amsmath}
\usepackage{amsfonts}
\usepackage{tabularx}
\usepackage{multirow}
\usepackage[caption=false,font=normalsize,labelfont=sf,textfont=sf]{subfig}
\usepackage{amsthm}
\usepackage{color}

\DeclareMathOperator*{\argmin}{arg\,min}
\DeclareMathOperator*{\argmax}{arg\,max}

\newtheorem{lemma}{Lemma}

\begin{document}

\title{Image Restoration Using Conditional Random Fields and Scale Mixtures of Gaussians}

\author{Milad Niknejad, Jos\'e  Bioucas-Dias, M\'ario A.T. Figueiredo
\thanks{}
\thanks{}
\thanks{The research leading to these results has received funding from the European Union's 
Seventh Framework Programme (FP7-PEOPLE-2013-ITN) under grant agreement n° 607290 SpaRTaN.}}

\markboth{}%
{Niknejad \MakeLowercase{\textit{et al.}}: Image Restoration Using Conditional Random Fields and Scale Mixtures of Gaussians}

\maketitle

\begin{abstract}
This paper proposes a general framework for internal patch-based image restoration based on \textit{conditional random fields} (CRF). Unlike related models based on \textit{Markov random fields} (MRF), our approach explicitly formulates the posterior distribution for the entire image. The potential functions are taken as proportional to the product of a likelihood and prior for each patch. By assuming identical parameters for similar patches, our approach can be classified as a model-based \textit{non-local} method. For the prior term in the potential function of the CRF model, multivariate Gaussians and multivariate scale-mixture of Gaussians are considered, with  the latter being a novel prior for image patches. Our results show that the proposed approach outperforms methods based on Gaussian mixture models for image denoising and state-of-the-art methods for image interpolation/inpainting.
\end{abstract}

% \fi
\IEEEpeerreviewmaketitle

\section{Introduction}
Restoring images from degraded versions thereof (\textit{e.g.}, noisy or blurred) is one of the fundamental and classical image processing problems, with a long history and countless applications. In recent years, several state-of-the-art image restoration methods have been proposed by exploiting the \textit{patch-based} paradigm. In most of those methods, the image is decomposed into small overlapping blocks, the \textit{patches}, which are processed (\textit{e.g.}, denoised) with the help of other patches  in the observed image itself (internal methods) \cite{2007_dabov_bm3d, 2013_dong_lssc,dong_nonlocally_2013,dong_image_2015}, or of a set of external clean patches (external methods) \cite{2005_roth_fields,2011_zoran_learning}, or a combination of both \cite{2016_luo_adaptive,2013_mosseri_combining}; finally, the processed patches are returned to their original location in the image, with the multiple estimates of overlapping pixels being somehow combined (typically averaged) to yield the whole restored image. 

In many approaches, the patch estimates are treated independently and are averaged (possibly in a weighted fashion) to construct the entire image estimate \cite{2009_dabov_sabm3d,2013_dong_lssc,Niknejad_TIP2015,xiong_image_2016}. Other authors consider the whole image model in their algorithms, which often (implicitly or explicitly) implies a way to combine the overlapping patch estimates \cite{2006_elad_denoisesparse,Figueiredo_ICASSP2017,2011_zoran_learning}.
The \textit{field of experts} (FoE) \cite{2005_roth_fields} is a patch-based probabilistic model for the entire image, which can be seen as \textit{Markov random field} (MRF) where the patches are the cliques and the potential functions are modeled using a \textit{product of experts} (PoE) \cite{2002_hinton_training}. The EPLL (\textit{expected patch log-likelihood} \cite{2011_zoran_learning}) method  also corresponds to a prior for the entire image that takes the form of a PoE (sum of log-potentials). Both the FoE and EPLL methods are usually applied as external restoration algorithms, meaning that the corresponding priors/models are learned from an external dataset of clean patches of natural images. 
%Beside applying entire image models, recently some methods additionally try to add some structures on the whole model such as convolutional sparse coding \cite{bristow_fast_2013}.

%Markov random fields (MRF) in the computer vision community, has been used to model the joint distribution of high dimensional random variables with special kind of dependencies. The clique is defined by a set of mutually connected random variables. Based on MRF, the joint probability can be expressed as the product of functions of cliques (called sometimes partition function) divided by the normalization factor. 

A common assumption of probabilistic patch-based models/priors is that each patch is independently drawn from a multivariate probability distribution. The distributions proposed so far in the context of image restoration have been essentially limited to Gaussian mixture models (GMM) \cite{2015_teodoro_single, 2012_yu_solving,2011_zoran_learning}, \textit{Gaussian scale mixtures} (GSM) \cite{Lyu_2006,2003_portilla_image},  or a set of multivariate normal distributions in which each of them is assumed to describe a set of similar patches \cite{lebrun_nonlocal_2013,Niknejad_TIP2015}. An important feature of GMM or Gaussian priors is that the \textit{minimum mean square error} (MMSE) patch estimates can be obtained in closed form \cite{2015_teodoro_single}. Although non-Bayesian methods have also been proposed (namely by exploiting the low-rank property of a stack of similar patches \cite{2013_dong_lssc}), probabilistic methods have some important advantages. Namely, they can be easily used in internal, external, or hybrid mode, allowing, for example, to learn class-specific models \cite{2015_luo_adaptive,2016_teodoro_image}. GMMs have been recently used to learn a combined internal and external prior \cite{2016_luo_adaptive}.  Some researchers have shown that natural image patches follow distributions with tails that are heavier that those of Gaussian densities \cite{2015_gerhard_modeling, deledalle_image_2018}; however, those models raise computational difficulties when used as priors to obtain Bayesian patch estimates. Recently, an \textit{importance sampling} \cite{Robert_Casela_1999} approach has been proposed to approximate MMSE estimation without any Gaussianity assumption \cite{niknejad_class_2017,niknejad_classp_2017}.

The probabilistic methods described above (namely those based on PoE, EPLL, or GMM) follow a \textit{generative} approach: they are based on models of the underlying clean image (\textit{i.e.}, \textit{priors}, in Bayesian terms), which are then combined with an observation model (\textit{e.g.}, the additive contamination with Gaussian noise) to yield a posterior with which inference can be carried out, \textit{i.e.}, image estimates can be obtained. The non-generative  alternative to probabilistic inference is based on directly modeling and using the posterior probability density of the unknown image, given the observations; a well-known instance of this approach uses the so-called \textit{conditional random fields} (CRF), introduced by Lafferty \textit{et al}  \cite{2001_lafferty_conditional}. CRF models have been widely used in many applications \cite{Sutton_CRF_2011}, namely in image analysis and computer vision \cite{2004_kumar_discriminative}, language processing \cite{2003_sha_shallow}, and bioinformatics \cite{2005_sato_rna}, to mention only a few well-known examples.

In this paper, we propose a general  patch-based  framework for image restoration using CRF models. Our method offers a posterior probability density function for the entire image, rather than considering independent patches followed by averaging to reconstruct the whole image. More specifically, we propose a new generic framework based on the modeling the entire image posterior distribution using a CRF model, where the cliques are the image patches. The experimental results reported in the paper show that our method outperforms the state-of-the-art in image inpainting and achieves competitive results in image denoising. 

In the following sections, first, the basics of CRF model is breifly mentioned, then, our proposed framework for image restoration is described. Two patch priors which are integrated to our model is, then, disscussed in details. Finally, the results of our proposed method compared to other methods are reported. 

%first, related methods to our work are discussed. Then,  our algorithm is explained. The two suggested priors of multivariate Gaussian and scale mixture of Gaussian are discussed in two subsections.  In the results' section, we evaluate our proposed method performance by comparing to other methods.

\section{Basic Tools  and Related Methods}
\label{sec:related}
Consider an undirected graph $G=(V,E)$ with a finite set $V$ of nodes and a set $E \subseteq V \times V$ of edges. Let $\mathcal{C} \subseteq 2^V$ be the set of (maximal) cliques\footnote{Recall that a subset of nodes $C \in 2^V$ is a \textit{clique} if any two of its elements are connected by an edge, \textit{i.e.}, if $\forall x,x' \in C, (x,x') \in E$. A maximal clique is a clique that is not a proper subset of any other clique.}.  A collection of random variables ${\bf X} = (X_v)_{v\in V}$, indexed by the elements of $V$, constitutes a \textit{Markov random field} (MRF) with respect to $G$ if the joint probability (density or mass) function can be written as
\begin{equation}
\label{eq:mrf}
p(\mathbf{x}) = \frac{1}{Z(\Theta)} \prod_{C \in \mathcal{C}} \phi_C (\mathbf{x}_C|\Theta_C),
\end{equation}
where ${\bf x}$ denotes a particular configuration of ${\bf X}$, $\mathbf{x}_C$ is the subset of elements of ${\bf x}$  in clique $C$, each function $\phi_C$ is called the potential of clique $C$ and has parameters $\Theta_C$ (with $\Theta$ denoting the full set of parameters of all the cliques), and $Z(\Theta)$ is the normalization constant, often called \textit{partition function} \cite{koller_probabilistic_2009}. Following the seminal work of  Geman and Geman \cite{Geman1984}, MRF models have been widely used in many areas, specially in computer vision, and image analysis/processing, usually as priors to be combined with an observation model to obtain Bayesian approaches to problems such as image restoration or segmentation. 

The \textit{field of experts} (FoE) model introduced by Roth \textit{et al} is a particular type of MRF model/prior for image pixels  \cite{2005_roth_fields}, where each clique is a local patch of the image to be restored, and the clique potentials all have the same form and are themselves products of non-negative functions, the so-called \textit{products of experts} (PoE), 
\[
\phi_C (\mathbf{x}_C|\Theta_C) = \prod_j \psi_j ({\bf J}_j^T {\bf x}_C,\Theta_j)
\]
where ${\bf x}_C$ is the vector holding all the pixels of clique/patch $C$ and ${\bf J}_j$ is the \textit{filter} corresponding to the $j$-th expert in the PoE.

The EPLL (\textit{expected patch log-likelihood} \cite{2011_zoran_learning}) formulation, although not originally introduced as a prior, does correspond to a prior with the form
\begin{equation}
\label{eq:epll}
p(\mathbf{x})=\prod_{i=1}^{m} p(\mathbf{R}_i\mathbf{x},\Theta),
\end{equation}
where $\mathbf{R}_i$ is a binary matrix that extracts the $i$-th patch from the image $\mathbf{x}$, and $p$ is some probability function. Clearly, \eqref{eq:epll} is an instance of \eqref{eq:mrf}, where each patch $\mathbf{R}_i \mathbf{x}$ is a clique and $\phi_C = p$, for all the cliques. Although the clique potentials in \eqref{eq:mrf} do not have to be (in general, are not) probability functions, this is a valid choice. Letting $p$ be a GMM has been shown to yield very good denoising results  \cite{2011_zoran_learning}. A variant of the EPLL which uses a sparse representation of each patch on a learned dictionary has also been proposed \cite{2015_sulam_expected}. Although for some choices of $p$, such as a Gaussian, the EPLL prior in (\ref{eq:epll}) can be written as an FoE, for many other choices, namely GMM (as used in  \cite{2011_zoran_learning}), this is not the case. 

\textit{Conditional random field} (CRF) models  are similar to MRF models \cite{kumar_discriminative_2003,2001_lafferty_conditional,Sutton_CRF_2011}, but the MRF construction is used to directly write the posterior probability (density of mass) function of the random variables to be estimated ${\bf X}$, given the observations ${\bf y}$, rather than a prior. With the same definitions as in the previous paragraphs, a CRF has the form
\begin{equation}
\label{eq:crf}
p(\mathbf{x}|\mathbf{y})=\frac{1}{Z(\Theta,\mathbf{y})} \prod_{C \in \mathcal{C}} \phi_C(\mathbf{x}_C|\mathbf{y},\Theta_C).
\end{equation}

Tappen \textit{et al} \cite{tappen_learning_2007} proposed an image denoising method  based on a \textit{Gaussian CRF} (GCRF). The potential function has an especial form in this case, which depends on a collection of convolution kernels  learned from an external dataset of clean images. Using this type of potential function in other image restoration problems is not straightforward.

%To address an image restoration problem with direct model ${\bf y = H x + n}$, where ${\bf H}$ is the observation operator (\textit{e.g.}, blur) and ${\bf n}$ is white Gaussian noise with variance $1/\gamma$, EPLL solves the minimization problem of the form
%\begin{equation}
%\label{eq:epll2}
%\hat{\bf x} \in \argmin_\mathbf{x} \frac{\gamma}{2} ||\mathbf{y}-\mathbf{H}\mathbf{x}||_2^2- \sum_i \log~p(\mathbf{R}_i\mathbf{x},\Theta),
%\end{equation}
%which corresponds to the MAP estimate, \textit{i.e.}, 
%\begin{equation}
%\hat{\bf x} = \arg\max_{\bf x} p({\bf x}|{\bf y}) = \arg\max_{\bf x} p({\bf y}|{\bf x})\; p({\bf x}).
%\end{equation}
%In order to solve \eqref{eq:epll2}, Zoran \textit{et al} \cite{2011_zoran_learning} used a variable splitting approach,
%\begin{equation}
%\label{eq:epll3}
%\argmin_{\mathbf{x},\mathbf{z}_i} \frac{\gamma}{2} ||\mathbf{y}-\mathbf{H}\mathbf{x}||_2^2- \sum_i \Bigl[ \frac{\beta}{2}||\mathbf{z}_i-\mathbf{R}_i\mathbf{x}||_2^2 - \log p(\mathbf{z}_i)\Bigr],
%\end{equation}
%which becomes equivalent to \eqref{eq:epll2} as $\beta$ goes to infinity.  A block-coordinate descent method is used to solve \eqref{eq:epll3}; the update of each patch ${\bf z}_i$ corresponds to a denoising problem, while updating ${\bf x}$ corresponds to fusing the current patch estimates to obtain a whole image estimate.

The  methods just described are all \textit{external}, in the sense that the corresponding parameters $\Theta$ are learned from an external dataset of clean images (or patches) and they only use one global prior distribution to model all the patches in the image to be estimated.

\section{Proposed Formulation}
\subsection{Estimation Criterion}
As mentioned above, our proposal is to use a CRF formulation, \textit{i.e.}, to directly model the posterior $p({\bf x}|{\bf y})$, rather than a prior $p({\bf x})$. More specifically, the posterior takes the form
\begin{equation}
\label{eq:factor}
p(\mathbf{x}|\mathbf{y}) = \frac{1}{Z(\Theta)}\prod_{i=1}^{m}{\phi (\mathbf{R}_i{\mathbf{x}}|\mathbf{y},\theta_i}) ,
\end{equation}
where $\mathbf{R}_i$ is a binary matrix that extracts the $i$-th patch from the image $\mathbf{x}$ and  $p(\cdot|{\bf y})$ is a patch-wise multivariate conditional probability density function. Notice that this is different from standard FoE or EPLL formulations, because we are directly specifying  the posterior, rather than a prior. Unlike in the EPLL and FoE formulations, the parameters $\theta_i$ of these conditional probability functions are not the same for all patches, this being the reason why we explicitly include them in the notation.

In denoising and inpainting problems, it is natural to assume that the pixels in each clean patch are conditionally independent from the observed pixels outside that patch, given the observations in the patch; formally, this means that  $\phi(\mathbf{R}_i{\mathbf{x}}|\mathbf{y},\theta_i )=\phi(\mathbf{R}_i{\mathbf{x}}|\mathbf{R}_i\mathbf{y},\theta_i )$. For deblurring problems, dependency  exists with respect to a larger window around the clean patch, denoted as $\mathbf{Q}_i\mathbf{y}$. In the sequel, we will denote $\mathbf{R}_i\mathbf{y}$ and $\mathbf{Q}_i\mathbf{y}$ simply as $\mathbf{y}_i$. Consequently, under the MAP criterion, the image estimate is given by
\begin{equation}
\label{eq:proposedMAP}
\hat{\mathbf{x}}=\argmax_\mathbf{x} \frac{1}{Z(\Theta)} \prod_{i=1}^{m} \phi(\mathbf{R}_i\mathbf{x}|\mathbf{y}_i,\theta_i ).
\end{equation}

%Our model further assumes that similar patches share the same parameters, \textit{i.e.}, $\mathbf{\theta}_i=\mathbf{\theta}_j$, if $d(\mathbf{R}_i\mathbf{x},\mathbf{R}\mathbf{x}_j)<t$ where $d$ is a distance measure. Imposing this assumption is inspired by successful patch-based methods such as BM3D \cite{2007_dabov_bm3d} which use collaborative filtering for similar patches in the image. The distance $d$, similar to these approaches, are considered as $l_2$ norm distance.

%In the case of multivariate Gaussian prior, there are two main differences in our model with the FOE method \cite{2005_roth_fields} for the image restoration: (a)  Our method is based on the CRF rather than MRF and (b) we use different potential function for along the image graph and the same only for similar patches. On the contrary to both FOE and EPLL, our method is completely internal and uses only the information of the noisy image.

\subsection{Instantiating the Model}
To instantiate the model, we need to define the form of the posterior potential functions $\phi(\mathbf{R}_i\mathbf{x}|\mathbf{y}_i,\theta_i)$, which, to keep the notation simple, we will denote as  $\phi(\mathbf{x}_i|\mathbf{y}_i,\theta_i)$, \textit{i.e.}, $\mathbf{x}_i$ denotes the $i$-th patch of ${\bf x}$. Our proposal is to let $\phi(\mathbf{x}_i|\mathbf{y}_i,\theta_i )$ have the same form as if we were estimating it alone from the corresponding observations ${\bf y}_i$, under some prior $p({\bf x}_i|\theta_i)$. That is, we write
\begin{equation}
\label{eq:mpotfunc}
 \phi(\mathbf{x}_i|\mathbf{y}_i) =  p(\mathbf{y}_i|\mathbf{x}_i) \; p(\mathbf{x}_i|\theta_i ),
\end{equation}
which,  plugged back into \eqref{eq:proposedMAP}, and after taking logarithms, leads to 
\begin{equation}
\label{eq:mainpr}
\hat{\mathbf{x}} =  \argmax_\mathbf{x} \sum_{i=1}^{m}\bigl( \log p(\mathbf{y}_i|\mathbf{R}_i\mathbf{x}) + \log p(\mathbf{R}_i\mathbf{x} | \theta_i)\bigr) .
\end{equation}
Our choice of potential function in \eqref{eq:mpotfunc}, unlike \cite{tappen_learning_2007}, readily allows extension to general image restoration problems.

Here, we consider only the classical linear-Gaussian observation model, where the observed image results from a linear operator applied to the original one, followed by the addition of white Gaussian noise of variance $\sigma^2$. In this case, (\ref{eq:mainpr}) can be written as
\begin{equation}
\label{eq:mainopt}
\hat{\mathbf{x}} = \argmin_\mathbf{x} \sum_{i=1}^{m}\bigl( \|\mathbf{y}_i-\mathbf{H}_i\mathbf{R}_i\mathbf{x}\|_2^2 - 2\sigma^2 \log p(\mathbf{R}_i\mathbf{x}|\theta_i)\bigr).
\end{equation}

Notice that, in the case of denoising, $\mathbf{H}_i = \mathbf{I}$. For inpainting problems, each $\mathbf{H}_i$ is a binary diagonal matrix, with the zeros in the diagonal corresponding to the pixels lost in that particular patch.

Below, we will address different choices of prior $p(\cdot |\theta)$ and  how to estimate the corresponding parameters. Before that, we will consider the problem of solving the optimization problem \eqref{eq:mainopt}, assuming that this prior is known.

\subsection{Split-and-Penalize}
\subsubsection{General Case}
In order to solve the  minimization problem in \eqref{eq:mainopt}, we use the \textit{split-and-penalize} approach (also known as \textit{half quadratic splitting} \cite{2011_zoran_learning}); \textit{i.e.}, we consider the equivalent constrained problem 
\begin{eqnarray}
\min_{\mathbf{x},\mathbf{z},\mathbf{q}} & & \sum_{i=1}^{m}\Bigl( \frac{\|\mathbf{y}_i-\mathbf{q}_i\|_2^2}{2\sigma^2}-\log p(\mathbf{z}_i; \theta_i)\Bigr) \label{eq:split1}\\
\mbox{subject to} & &   \mathbf{z}_i=\mathbf{R}_i\mathbf{x}, \;\; i=1,...,m\nonumber\\
& & \mathbf{q}_i=\mathbf{H}_i\mathbf{z}_i, \;\; i=1,...,m\nonumber,
\end{eqnarray}
and add penalty terms 
\begin{eqnarray}
\min_{\mathbf{x},\mathbf{z},\mathbf{q}} \sum_{i=1}^{m}\Bigl(\frac{\|\mathbf{y}_i-\mathbf{q}_i\|_2^2}{2\sigma^2}-\log p(\mathbf{z}_i; \theta_i)+\frac{\lambda}{2}\|\mathbf{R}_i\mathbf{x}-\mathbf{z}_i\|_2^2 \nonumber \\ +\frac{\rho}{2}\|\mathbf{H}_i\mathbf{z}_i-\mathbf{q}_i\|_2^2\Bigr),
\label{eq:varsplit}
\end{eqnarray}
where $\lambda \geq 0$ and $\rho \geq 0$ are the so-called penalty parameters \cite{courant1943variational}.

The key observation is that problem \eqref{eq:varsplit}, as both $\lambda$ and $\rho$ approach infinity, becomes equivalent to \eqref{eq:split1}. This suggests that a natural approach is to use a block-coordinate descent algorithm (\textit{i.e.}, minimizing \eqref{eq:varsplit} in turn with respect to $\mathbf{x}$, $\mathbf{z}$, and $\mathbf{q}$, while keeping the other two at their current values), while increasing $\lambda$ and $\rho$.

Solving \eqref{eq:varsplit} with respect to the $\mathbf{q}_i$'s, with the other variables fixed, leads to $m$ independent problems
\begin{equation}
\mathbf{q}_i^{\mbox{\tiny new}} = \argmin_{\mathbf{q}_i} \frac{\|\mathbf{y}_i-\mathbf{q}_i\|_2^2}{2\sigma^2}+\frac{\rho}{2}\|\mathbf{H}_i\mathbf{z}_i-\mathbf{q}_i\|_2^2,
\end{equation}
for $i=1,\ldots,m$, which have simple closed-form solutions
\begin{equation}
\label{eq:updateq}
\mathbf{q}_i^{\mbox{\tiny new}}=\frac{\mathbf{y}_i+\sigma^2\,\rho\, \mathbf{H}_i\mathbf{z}_i}{1+\sigma^2\rho},\, \qquad \mbox{for}\; i=1,\ldots,m.
\end{equation}

Solving (\ref{eq:varsplit}) with respect to $\mathbf{x}$, while the other variables are kept fixed, leads to the solution 
\begin{equation}\label{eq:update_x}
\mathbf{x}^{\mbox{\tiny new}}= \biggl( \sum_{i=1}^m\mathbf{R}_i^{T}\mathbf{R}_i \biggr)^{-1} \sum_{i=1}^m \mathbf{R}_i^T\mathbf{z}_i;
\end{equation}
from the image point of view, this corresponds to returning each $\mathbf{z}_i$ (which has the size of a patch) to the position of the $i$-th patch in the image, averaging overlapping entries \cite{2011_zoran_learning}.

Finally, solving with respect to the $\mathbf{z}_i$'s depends on the prior adopted, but is separable across the $\mathbf{z}_i$. These optimization problems can be written compactly as 
\begin{equation}
\mathbf{z}_i^{\mbox{\tiny new}}= \argmin_{\mathbf{z}_i}  -\log p(\mathbf{z}_i; \theta_i)+\frac{\lambda}{2}\|\mathbf{R}_i\mathbf{x}-\mathbf{z}_i\|_2^2+\frac{\rho}{2}\|\mathbf{H}_i\mathbf{z}_i-\mathbf{q}_i\|_2^2  ,\label{eq:zinew}
\end{equation}
for $i=1,\ldots,m$. Addressing these optimization problems will be deferred to the next subsection, since it depends on the choice of the prior.

\subsubsection{Denoising Case}
In the case of denoising problems, $\mathbf{H}_i = \mathbf{I}$, which implies that the problem formulation can be simplified. Instead of \eqref{eq:split1}, we only need to consider
\begin{eqnarray}
\min_{\mathbf{x},\mathbf{z}}, & & \sum_{i=1}^{m}\Bigl( \frac{\|\mathbf{y}_i-\mathbf{z}_i\|_2^2}{2\sigma^2}-\log p(\mathbf{z}_i; \theta_i)\Bigr) \label{eq:split_denoise}\\
\mbox{subject to} & &   \mathbf{z}_i=\mathbf{R}_i\mathbf{x}, \;\; i=1,...,m\nonumber
\end{eqnarray}
and the corresponding penalized unconstrained problem becomes
\begin{equation}
\label{eq:varsplit_den}
\min_{\mathbf{x},\mathbf{z}} \sum_{i=1}^{m}\Bigl(\frac{\|\mathbf{y}_i-\mathbf{z}_i\|_2^2}{2\sigma^2}-\log p(\mathbf{z}_i; \theta_i)+\frac{\lambda}{2}\|\mathbf{R}_i\mathbf{x}-\mathbf{z}_i\|_2^2\Bigr).
\end{equation}
The update equation for $\mathbf{x}$ is still \eqref{eq:update_x}, variables $\mathbf{q}_i$ do not exist, and the update of the $\mathbf{z}_i$'s is done by minimizing \eqref{eq:varsplit_den} with respect  to these variables, which can be done independently for each one of them.

\section{Patch Priors}
In this paper, we consider two types of patch priors: Gaussian and scale-mixture of Gaussians. We show how these choices reflect on the steps of the algorithm outlined above, and present the proposed strategy for estimating the corresponding parameters, yielding the final complete algorithms.

\subsection{Gaussian Priors}

\subsubsection{Updating $\mathbf{z}_i$}
We begin by assuming that the parameters $\theta_i$ are known; later we will describe the approach herein proposed to estimate these parameters. By adopting a Gaussian prior for each patch $i$, with mean  $\mbox{\boldmath$\mu$}_i $ and covariance $\mathbf{C}_i$, we have 
\begin{equation}
-\log p(\mathbf{z}_i; \theta_i) = \frac{1}{2}(\mathbf{z}_i - \mbox{\boldmath$\mu$}_i)^T \mathbf{C}_i^{-1} (\mathbf{z}_i - \mbox{\boldmath$\mu$}_i)+c,
\end{equation}
where $\theta_i = (\mbox{\boldmath$\mu$}_i,\mathbf{C}_i )$ and $c$ is a constant. In this case,  \eqref{eq:zinew} has a simple closed-form solution
\begin{equation}
\label{eq:gest}
\mathbf{z}_i^{\mbox{\tiny new}} = \Bigl( \mathbf{C}_i^{-1} + \lambda \mathbf{I}+\rho \mathbf{H}_i^T\mathbf{H}_i \Bigr)^{-1} \Bigl(
\mathbf{C}_{i}^{-1}  \mbox{\boldmath$\mu$}_i  +\lambda\mathbf{R}_i\mathbf{x}+\rho\mathbf{H}_i^T\mathbf{q}_i \Bigr),
\end{equation}
for $i=1,\ldots,m$. For the image denoising case, this step is slightly different, as it corresponds to solving \eqref{eq:varsplit_den} with respect to each $\mathbf{z}_i$, which leads to
\begin{equation}
\mathbf{z}_i^{\mbox{\tiny new}} = \Bigl( \mathbf{C}_i^{-1} + \bigl(\lambda + (1/\sigma^2)\bigr) \mathbf{I} \Bigr)^{-1} \Bigl(
\mathbf{C}_{i}^{-1}  \mbox{\boldmath$\mu$}_i  + \lambda\mathbf{R}_i\mathbf{x} + \mathbf{y}_i /\sigma^2 \Bigr).
\end{equation}

\subsubsection{Parameter Update}
In all related methods based on MRF and CRF discussed in Section \ref{sec:related}, the parameters of the potential functions are learned using the external dataset of clean images. However, in our proposed method, the availability of an external dataset is not assumed as it is the case in many applications. We assume that similar patches share similar parameters. This assumption is used in many patch-based image restoration method based on collaborative filtering. Three related examples are \textit{piece-wise linear estimator} (PLE) \cite{2012_yu_solving}, \textit{linear estimator with neighborhood clustering} (LINC) \cite{Niknejad_TIP2015} and \textit{non-local Bayes} (NL-Bayes) \cite{lebrun_nonlocal_2013}, which assume that similar patches are described by a common multivariate Gaussian distribution. Here, we aim to obtain the parameters for $\mathbf{z}_i$'s which are actually image patches before averaging in \eqref{eq:zinew}.  The procedure of selection of similar patches $\mathbf{z}_i$ in our method is similar to the clustering in the BM3D algorithm, and will be discussed in details in implementation section. For now, assume a set of similar patches denoted by $\mathcal{I}_c$ is available at hand. Following the block coordinate descent algorithm in our method,  the parameters are estimated using the same procedure. {Keeping $\mathbf{z}_i$'s fixed, \eqref{eq:varsplit_den} is solved with the respect to $\theta_i$ based on the constraint of the same covariance matrix for similar patches. This corresponds to the Maximum Likelihood estimation which leads to the sample mean and the sample covariance matrix \textit{i.e.} for patch $i \in \mathcal{I}_c$}
\begin{equation}
\label{eq:smplcov}
\hat{\boldsymbol{\mu}}_i=\frac{1}{n} \sum_{i \in \mathcal{I}_c}{\mathbf{z}_i}, \ \hat{\mathbf{C}}_i=\frac{1}{n} \sum_{i \in \mathcal{I}_c}{(\mathbf{z}_i-\hat{\boldsymbol{\mu}}_i)(\mathbf{z}_i-\hat{\boldsymbol{\mu}}_i)^T}.
\end{equation}

\subsection{Gaussian Scale Mixture Prior}

Some research \cite{deledalle_image_2018,2015_gerhard_modeling} has shown that the patch statistics follow higher kurtosis (heaver tails and higher peaks) distributions compared to Gaussian. However, due to difficulty of obtaining estimates using these priors, the patch priors have been mostly limited to multivariate Gaussians or GMM. In this section, we consider \textit{Gaussian scale mixture} (GSM) priors, which generalize the multivariate Gaussian and can be adapted to model higher kurtosis distributions.

\subsubsection{Updating $\mathbf{z}_i$}
The random variable follows a GSM distribution if it can be written as a product of two independent random variables as
\begin{equation}
\label{eq:smgprod}
\mathbf{z} = \sqrt{v}\; \mathbf{u},
\end{equation}
where $\mathbf{u}$ follows multivariate Gaussian, and $v\in \mathbb{R}_+$ is a non-negative random variable, which is independent of $\mathbf{u}$.
A GSM has the important property that the conditional distribution of $\mathbf{z}$ given $v$ is simply a multivariate Gaussian with the mean $\sqrt{{v}}\, \boldsymbol{\mu}_{u}$ and covariance ${v}\mathbf{\Sigma}$, where $\boldsymbol{\mu}_{u}$ and $\mathbf{\Sigma}$ are the mean and covariance of $\mathbf{u}$, respectively. 

Let us return to the problem of updating the estimate of each $\mathbf{z}_i$ in \eqref{eq:zinew}, now with $p$ corresponding to a GSM prior. For a moment, assume the value of the underlying ${v}_i$ is known. Given ${{v}}_i$, the prior becomes Gaussian, thus
\begin{eqnarray}
\mathbf{z}_i^{\mbox{\tiny new}}=({v}_i^{-1}\mathbf{\Sigma}_{i}^{-1}+\lambda \mathbf{I}+\rho \mathbf{H}_i^T\mathbf{H}_i)^{-1}({{v}_i}^{-\frac{1}{2}}\mathbf{\Sigma}_{i}^{-1}\boldsymbol{\mu}_{i}+\lambda\mathbf{R}_i\mathbf{x} \nonumber \\ +\rho\mathbf{H}_i^T\mathbf{q}_i).
\label{eq:sgest}
\end{eqnarray}

Updating the variable $\mathbf{z}_i$ in above requires an estimation of the latent random variable ${v}_i$ as well as the parameters of the underlying Gaussian. Here, we discuss how to obtain an estimate of ${v}_i$, and defer the problem of parameter updating to the next subsection.

Considering the independence of  $\mathbf{u}_i$ and $v_i$, the mean and covariance of $\mathbf{z}_i$ are $\boldsymbol{\mu}_{z_i}=\mathbb{E}(\sqrt{v}_i) \boldsymbol{\mu}_{u_i}$ and $\mathbf{C}_i=\mathbb{E}(v_i)\mathbb{E}(\mathbf{u}_i {\mathbf{u}_i}^T) - {\mathbb{E}^2(\sqrt{v_i})}\boldsymbol{\mu}_{u_i} \boldsymbol{\mu}_{u_i}^T$, respectively.
To obtain an estimate of $v_i$, we use the MAP criterion, conditioned on the current estimate of $\mathbf{z}_i$. The prior for $v_i$ is considered to be a Gamma distribution which will be shown to lead to a tractable solution. Assuming the current value of $\mathbf{z}_i$ is known, the MAP estimate is written as
\begin{equation}
{v}_i^{(new)}=\argmax_{v_i} {p(v_i|{\mathbf{z}}_i)}=\argmax_{v_i}{p({\mathbf{z}}_i|v_i)p(v_i)}.
\end{equation}
In the above, the first term is the multivariate normal distribution, and the second term is the prior distribution for $v_i$ which is a Gamma distribution with parameters $\alpha$ and $\beta$.
The above MAP estimate is then given by
\begin{equation}
\label{eq:zprop0}
{v}_i^{(new)}=\argmax_{v_i}{\frac{\beta^\alpha v_i^{\alpha-1}e^{-\beta v_i}}{\Gamma(\alpha)}} \frac{e^{-\frac{1}{2}\|\mathbf{z}_{i}-\sqrt{v_i} \boldsymbol{\mu}_{u_i}\|^2_{{(v_i\mathbf{\Sigma}_i)}^{-1}}}}{\sqrt{|v_i\mathbf{\Sigma}_i|}}
\end{equation}
where $\|\mathbf{z}-\boldsymbol{\mu}\|^2_{{\mathbf{W}}}={{(\mathbf{z}-\boldsymbol{\mu})}^T \mathbf{W}(\mathbf{z}-\boldsymbol{\mu})}$.
\begin{lemma}
{The solution to the above optimization problem is  the square root of the positive solution of the following $4^{th}$ order equation
\begin{equation}
\beta {w}_i^4+(1-\alpha+\frac{n}{2}){w}_i^2+\frac{c_i}{2} w_i-\frac{d_i}{2}=0
\label{eq:quartic}
\end{equation}
where $d_i={({\mathbf{z}}_{i})}^T \mathbf{\Sigma}_i^{-1}({\mathbf{z}}_{i})$, and $c_i={(\mathbf{z}_{i})}^T \mathbf{\Sigma}_i^{-1}(\boldsymbol{\mu}_{\mathbf{u}_i})$.}
\end{lemma}
\proof See the appendix.

{Based on above, our method requires the real positive solution for quartic equation in \eqref{eq:quartic}. Although the solutions of quartic equations are well-known, the analysis of guaranteeing a positive solution in \eqref{eq:quartic} is quite cumbersome. However, in our application, we found that the coeficients $c_i$ and $d_i$ in \eqref{eq:quartic} are much larger than the other coeficients \footnote{In our experiments, $c_i$ and $d_i$ are in the order of $10^9$.}. In this regime, the solution can be obtined by disregarding other coeficients (as they can be considered zeros) and compute $v_i$ as $\sqrt{\frac{d_i}{c_i}}$.} More percisely, as $c_i$ and $d_i$ increase, three roots of \eqref{eq:quartic} diverge to infinity while one is fixed at ${\frac{d_i}{c_i}}$.

\subsection{Parameter Update}
\label{sec:pupdate}
Estimating $v_i$ and $\mathbf{u}_i$ requires setting the parameters of the Gamma prior ($\alpha$ and $\beta$) and the covariance matrix of $\mathbf{u}_i$, denoted $\mathbf{\Sigma}_i$. However, only $\mathbf{C}_i$ (the covariance matrix of $\mathbf{z}_i$) can be estimated from the available samples of $\mathbf{z}_i$, by the sample covariance matrix. Using the following lemma, $\mathbf{\Sigma}_{i}$ can be obtained from $\mathbf{C}_{i}$, using a specific setting of the parameters of the Gamma prior.

\begin{lemma}
If the parameters of Gamma prior  satisfy $\beta=\Gamma(\alpha) \, \sqrt{\alpha}/{\Gamma(\alpha+\tfrac{1}{2})}$, the covariance matrix of the latent variable $\mathbf{u}_i$ satisfies $\mathbf{C}_i=(\alpha/\beta) \mathbf{\Sigma}_i$.
\end{lemma}
\proof See the appendix.

Similar to the multivariate Gaussian prior in \eqref{eq:smplcov}, the covariance matrix $\mathbf{C}_i$ is estimated as the sample covariance of similar patches. Based on the above lemma, by setting $\beta=\Gamma(\alpha) \, \sqrt{\alpha}/{\Gamma(\alpha+\tfrac{1}{2})}$, then the estimate of $\mathbf{\Sigma}_i$ is obtained by simply scaling that of $\mathbf{C}_i$ by the factor $\beta/\alpha$. Furthermore, $\alpha$ is hand-tuned to yield good restoration results. In our experiments, we set $\alpha=1/2$ and $\beta= \Gamma(1/2) / (\sqrt{2}\, \Gamma(1) )  \simeq 5/4$.

Though the whole procedure described for obtaining the estimate of $\hat{\mathbf{z}}_i$ under the GSM prior, seems computationally demanding, a closer look reveals that it roughly requires only the additional solution of the univariate quartic equation in \eqref{eq:quartic}, when compared to the Gaussian prior.  In summary, the algorithm to obtain the patch estimates has the following steps: 
\begin{description}
	\item[1.] estimate the covariance matrix $\mathbf{C}_i$ from the cluster of smilar patches;
	\item[2.]  compute the estimate of $\mathbf{\Sigma}_i$ by multiplying the estimate of  $\mathbf{C}_i$ by $\beta/\alpha$;
	\item[3.] {compute ${v}_i$ as $\sqrt{\frac{d_i}{c_i}}$;} 
	\item[4.] obtain the patch estimate according to \eqref{eq:sgest}.
\end{description}

It is worth to clarify the differences between our approach and the estimation procedure  in \cite{2003_portilla_image},  both based on a GSM prior.  In addition to the fact that \cite{2003_portilla_image} works on the wavelet domain, that work uses MMSE estimation; this involves computing an integral, which is approximated numerically by sampling from the scale parameter, resulting in a computationally expensive procedure.  Our approach only requires solving a univariate quadratic equation and the multiplication in \eqref{eq:sgest}. In \cite{2003_portilla_image},  unit mean is assumed for the scale variable, while in our method, the scale variable is continuous and the mean is the mean of the underlying Gamma distribution. Although, in the wavelet domain, the coefficients are zero-mean, the mean of grouped patches in our method are not zero in general.

\section{Practical implementation}
In this section, we describe the implementation details of our proposed method. As mentioned, we assume that similar patches have the same parameters in the potentials functions. The approach of determining similar patches in our method is similar to state-of-the-art patch-based methods, namely BM3D \cite{2007_dabov_bm3d}. In BM3D, a set of $k$ most similar (in Euclidean distance) patches to a reference patch is found within a neighborhood of that reference patch. This neighborhood is taken as a fixed-size window around the reference patch, and the reference patches are selected with fixed step sizes along the rows and columns of the image. In the BM3D patch grouping method, it often happens that one patch may belong to more than one cluster, while our method requires that each patch belongs to one and only one cluster, since one set of parameters is allowed for each patch. In order to adhere to our model, we consider the clustering used in BM3D and, in the cases where more than one estimate is obtained for each patch (a patch belongs to more than one cluster), one of them is chosen randomly.  We store the  previous versions of the vectors $\mathbf{q}_i$ and $\mathbf{z}_i$, which are required for the subsequent iteration, in two matrices $\mathbf{Q}$ and $\mathbf{Z}$, respectively.

The parameters $\lambda$ and $\rho$ of the algorithm (see \eqref{eq:varsplit}) should increase along the iterations. We empirically found that an exponential increase leads to good results, thus we update these parameters using $\lambda_{(l+1)}=\gamma_1 \lambda_{(l)}$ and $\rho_{(l+1)}=\gamma_2 \rho_{(l)}$ where $\gamma_1 > 1$ and $\gamma_2 > 1$.

Figure~\ref{fig:imgdenoise} shows the complete algorithm of our proposed method.
\begin{figure}[!h]
\centering
\begin{tabular}{|p{8.1cm}|}
\hline
\begin{itemize}
\setlength{\itemindent}{-1em}
\item Initialization: $\hat{\mathbf{X}}_0=\mathbf{Y}$, $\lambda=\lambda_0$, $\rho=\rho_0$, $\mathbf{Q}$ and $\mathbf{Z}$ with zero matrices, and $v_i=1$ for all the patches.
\item Main loop: for $l=1, \ldots ,L$
\begin{itemize}
\setlength{\itemindent}{-1em}
\item select reference patches along the row and the column of the image
\item For each reference patch:
\begin{itemize}
\setlength{\itemindent}{-1em}
\item Select $k$ nearest neighbor patches to the reference patch in a finite-size window.
\item Compute Gaussian parameters by \eqref{eq:smplcov}.
\item Obtain an estimation of restored patches $\mathbf{z}_i$'s by either \eqref{eq:gest} or the four step procedure in section \ref{sec:pupdate}, depending on the prior.
\item Obtain $\mathbf{q}_i$'s by \eqref{eq:updateq}.
\item Update $\mathbf{Q}$ and $\mathbf{Z}$ by the corresponding $\mathbf{q}_i$ and $\mathbf{z}_i$ values.
\end{itemize}
\item Obtain reconstructed image $\mathbf{X}_{(l)}$ by simple averaging of patches.
\item Update the regularization parameters by $\lambda_{(l+1)}=\gamma_1 \lambda_{(l)}$ and $\rho_{(l+1)}=\gamma_2 \rho_{(l)}$.
\end{itemize}
\item Final restored image is $\hat{\mathbf{X}}_{(L)}$.
\end{itemize}
 \\
\hline
\end{tabular}
\caption{The algorithm of the proposed image restoration method.}
\label{fig:imgdenoise}
\end{figure}

Our approach can be classified in a group of method sometimes called \textit{non-local} methods \cite{2007_dabov_bm3d,Niknejad_TIP2015,2014_gu_weighted}, in which patches are extracted from the image and then some kind of collaborative filtering is applied on the group of similar patches and then patches are returned to the original position in the image and are averaged in the overlapping pixels. Our CRF-based approach gives a formal support to this averaging procedure. The proposed model offers some modifications in the approach of the mentioned non-local methods, which result from using the CRF model. 

\section{Results}
This section reports results obtained in image denoising and inpainting experiments. The proposed method is compared to other internal methods. Among external methods, EPLL and FOE are also considered in our comparison, since their frameworks are closely related to our method.

\subsection{Image denoising}
As mentioned above, in denoising, the auxiliary parameter $\rho$ is not needed, and only $\alpha$ should be determined along the iterations. The initial value is set to $\lambda_0=10^{-4}$, and the incremental coefficient  $\gamma_1$ is set to $1.2$ for all noise levels. The parameters of the Gamma distribution for the scale parameter $v_i$ based on Lemma 1 were fixed to $\alpha=.5$ and $\beta=1.25$ for all noise levels. For all image restoration tasks in this paper, 39 similar patches (of size  $8\times8$) are selected for each reference patch in a window of size of $40\times40$. Reference patches are selected every 5 pixels along the row and column of the image. The results are reported for $10$ iterations of the block coordinate descent algorithm.
\begin{table*}[h!]
\caption{Image denoising PSNR (dB) Results}
\label{tab:denoise}
\renewcommand*{\arraystretch}{1.3}
\newcolumntype{C}[1]{>{\hsize=#1\hsize\centering\arraybackslash}X}%
\begin{tabularx}{\textwidth}{|C{}|C{}|C{}|C{}|C{}|C{}|C{}|C{}|C{}|C{}|C{}|C{}|C{}|}
\hline
&\multicolumn{6}{|c|}{$\sigma=10$}&\multicolumn{6}{|c|}{$\sigma=20$} \\
\hline
&BM3D&EPLL&LINC \cite{Niknejad_TIP2015}&WNNM&Ours (GMM)&Ours (SMG)&BM3D&EPLL&LINC \cite{Niknejad_TIP2015}&WNNM&Ours (GMM)&Ours (SMG)\\
\hline
Peppers&{34.68}&34.54&34.63&\bf{34.95}&{34.66}&{34.68}&31.29& 31.27&31.28&\bf{31.57}&31.35&{31.41} \\ 
House&36.71&35.75&36.76&\bf{36.95}&{36.72}&{36.85}&33.77&33.07&33.93&\bf{34.06}&33.97&\bf{34.05} \\
C.man&34.18&34.02&34.07&3\bf{4.44}&{34.09}&34.21&30.48&30.23&30.36&\bf{30.68}&30.41&{30.49} \\
Barbara&34.98&33.61&35.07&\bf{35.51}&{35.21}&35.36&31.78&29.77&32.04&31.66&32.10&\bf{32.17}\\
Lena&35.93&35.58&35.87&\bf{36.03}&{35.91}&\bf{36.02}&33.05&32.59&{33.09}&\bf{33.18}&33.05&{33.11} \\
Man&33.98&33.97&33.90&\bf{34.23}&{34.06}&{34.04}&{33.98}&30.57&33.90&\bf{33.91}&33.81&33.75 \\
\bf{Avg.}&35.08&34.58&35.05&\bf{35.35}&35.11&{35.18}&32.39&31.25&32.43&\bf{32.51}&32.45&{32.49} \\ \hline
&\multicolumn{6}{|c|}{$\sigma=30$}&\multicolumn{6}{|c|}{$\sigma=50$} \\ \hline
Peppers&29.28&29.16 &{29.46}&\bf{29.52}&29.34&{29.40}&26.68&26.63&{26.86}&\bf{26.91}&{26.71}&{26.82} \\
House&32.09&31.23 &32.23&\bf{32.59}&32.31&{32.42}&29.69&28.76&30.04&\bf{30.32}&{29.99}&{30.22} \\
C.man&{28.64}&28.36&28.44&\bf{28.78}&28.32&{28.52}&26.12&26.02&26.42&26.45&{26.29}&\bf{26.51} \\
Barbara&29.81&27.57&30.13&29.53&30.04&\bf{30.15}&27.23&26.82&27.42&\bf{27.79}&{27.37}&\bf{27.48} \\
Lena&31.26&30.79&{31.48}&\bf{31.52}&31.32&{31.34}&{29.05}&28.42&29.04&\bf{29.24}&{28.87}&28.94 \\
Man&{28.86}&28.83 &28.76&\bf{28.91}&28.75&{28.77}&\bf{26.81}&26.72&26.67&\bf{26.94}&{26.62}&26.66 \\
\bf{Avg.}&29.99&29.32&30.08&\bf{30.14}&30.05&{30.10}&27.59&27.22 &27.74&\bf{27.94}&27.62&{27.75} \\ \hline
\end{tabularx}
\end{table*}

\begin{figure*}[!h]
\centering
\subfloat[]{\includegraphics[width=.49\textwidth]{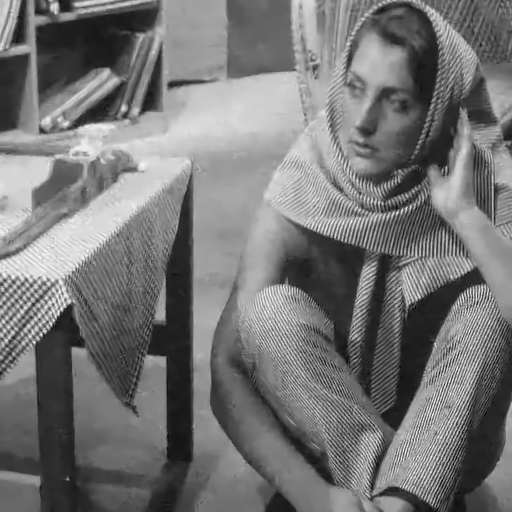}%
\label{(a)}}~
\subfloat[]{\includegraphics[width=.49\textwidth]{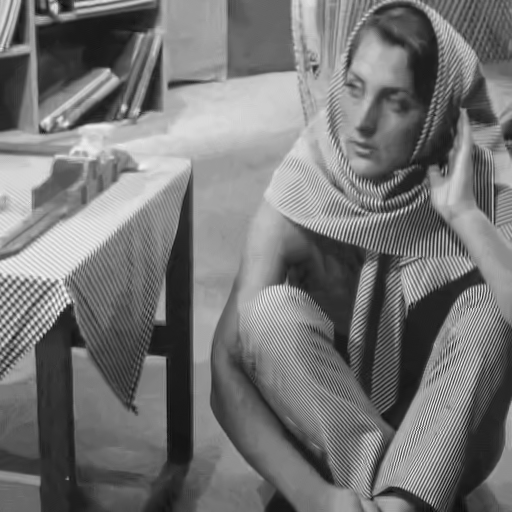}%
\label{(b)}}
\hfil
\subfloat[]{\includegraphics[width=.49\textwidth]{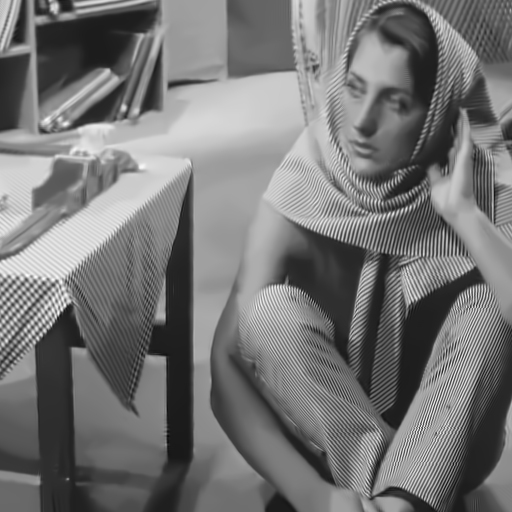}%
\label{(c)}}~
\subfloat[]{\includegraphics[width=.49\textwidth]{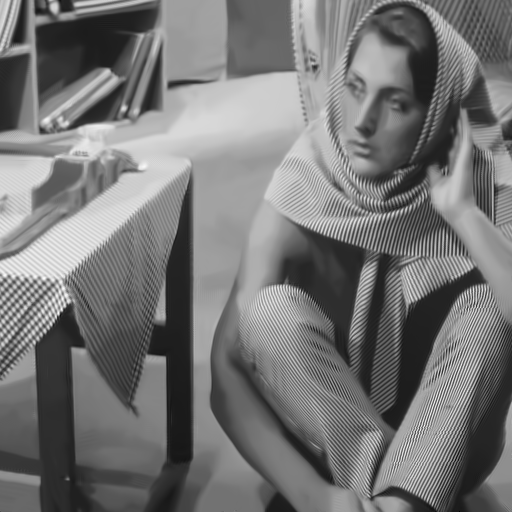}%
\label{(d)}}
\caption{Example of denoising results for the Barbara image at $\sigma=20$: (a) EPLL (PSNR=29.77 dB) (b) BM3D (PSNR=31.69 dB); (c) LINC (PSNR=32.04 dB); (d) This work (GSM) (PSNR=32.17 dB).}
\label{fig:barbara10}
\end{figure*}

In table \ref{tab:denoise}, PSNR values of images denoised by our method are compared to other methods:  BM3D, EPLL, LINC, and the current state-of-the-art internal image denoising method (\textit{weighted nuclear norm minimization}--WNNM--\cite{2014_gu_weighted}). It can be seen that our method outperforms other methods, on average PSNR, except WNNM. Our method yields results  comparable with WNNM on several images. It can also be concluded that the GSM prior performs better than the multivariate Gaussian prior.

An example of denoising result of our method for the Barbara image comparing to other methods is shown in Fig. \ref{fig:barbara10}.

\subsection{Image Inpainting}
In image inpainting, the goal is to estimate missing pixels. We consider both the noisy and noiseless cases.
In our implementation, the initial values of the parameters  $\lambda_{(0)}$ and $\rho_{(0)}$ were set to $10^{-6}$ and $.02$, respectively, and $\gamma_1$ and $\gamma_2$ were set to $1.35$ and $1.5$, respectively. Other parameters are the same as the Gaussian denoising case.   The initial covariance matrix and the mean vector are obtained as in \cite{Niknejad_TIP2015}, \textit{i.e.}, using only the observed pixels.

Table \ref{tab:intre} reports the results of our method compared to other methods for image interpolation with irregular missing data without additive noise. The results of our method are reported for both multivariate Gaussian and GSM priors. The random variables defining the missing pixels follow independent Bernoulli distributions with parameter $p$ corresponding to the percentage of missing data. Our method is compared to the \textit{Beta process} (BP) \cite{zhou_nonparametric_2012}, \textit{kernel regression} (KR) \cite{takeda_kernel_2007}, FoE \cite{2005_roth_fields}, and LINC \cite{Niknejad_TIP2015}.  Note that LINC, similarly to the proposed method, applies the same multivariate Gaussian prior to similar patches in a neighborhood. However, the method heuristically uses a weighted average of restored patches and lacks a global framework for representing the posterior distribution. The results show that our method outperforms all other  methods. It also shows a noticeable improvement over LINC, which can indicate the advantage of the CRF model on the image. The effectiveness of the GSM prior can also be seen in this table, as it improves upon the multivariate Gaussian prior in the same framework.

\begin{table*}[h!]
\caption{Image inpainting PSNR (dB) Results}
\label{tab:intre}
\renewcommand*{\arraystretch}{1.3}
\newcolumntype{C}[1]{>{\hsize=#1\hsize\centering\arraybackslash}X}%
\begin{tabularx}{\textwidth}{|C{1.0}|C{1.3}|C{.95}|C{.95}|C{.95}|C{.95}|C{.95}|C{.95}|}
\hline
&\% of available pixels&BP \cite{zhou_nonparametric_2012}&KR \cite{takeda_kernel_2007}&FOE \cite{2005_roth_fields}&LINC& This Work (GMM)&This Work (SMG) \\
\hline
\multirow{3}{*}{Barbara}&80\%&40.76&37.81&38.27&\bf{43.92}&43.28&{43.37} \\ \cline{2-8}
&50\%&33.17&27.98&29.47&{37.48}&37.79&\bf{38.01} \\ \cline{2-8}
&30\%&27.52&24.00&25.36&{33.68}&34.48&\bf{34.59} \\ \hline
\multirow{3}{*}{Lena}&80\%&41.27&41.68&42.17&\bf{43.60}&43.42&{43.51} \\ \cline{2-8}
&50\%&36.94&36.77&36.66&{37.98}&38.16&\bf{38.29} \\ \cline{2-8}
&30\%&33.31&33.55&33.22&{34.53}&34.90&\bf{35.08}\\ \hline
\multirow{3}{*}{House}&80\%&43.03&42.57&44.70&{45.21}&45.91&\bf{46.40} \\ \cline{2-8}
&50\%&38.02&36.82&37.99&{39.43}&40.10&\bf{40.22} \\ \cline{2-8}
&30\%&33.14&33.62&33.86&{35.16}&35.76&\bf{36.78} \\ \hline
\multirow{3}{*}{Boats}&80\%&39.50&37.91&38.33&{40.70}&40.76&\bf{40.82} \\ \cline{2-8}
&50\%&33.78&32.70&33.22&{34.58}&34.78&\bf{34.90} \\ \cline{2-8}
&30\%&30.00&29.28&29.80&30.81&30.90&\bf{31.01} \\ \hline
\end{tabularx}
\end{table*}

\begin{figure*}

\subfloat[]{\includegraphics[width=.32\textwidth]{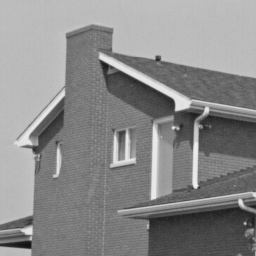}%
\label{(a)}}~
\subfloat[]{\includegraphics[width=.32\textwidth]{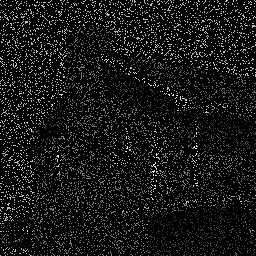}%
\label{(b)}}~
\subfloat[]{\includegraphics[width=.32\textwidth]{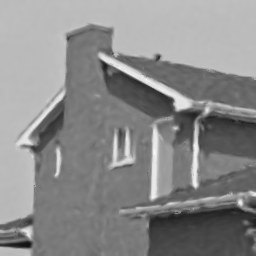}%
\label{(c)}}~
\\
\hfill
\subfloat[]{\includegraphics[width=.32\textwidth]{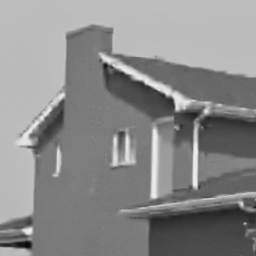}%
\label{(d)}}~
\subfloat[]{\includegraphics[width=.32\textwidth]{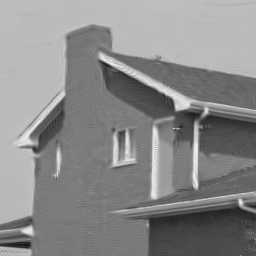}%
\label{(e)}}~
\subfloat[]{\includegraphics[width=.32\textwidth]{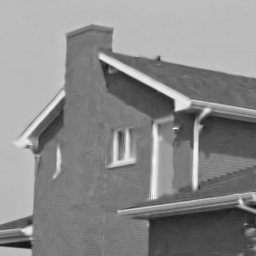}%
\label{(f)}}
\hfill
\caption{Example of image interpolation from irregular sample pixels. (a) Original image; (b) Degraded image with $15\%$ of pixels available; (c) KR (PSNR=29.65 dB); (d) EPLL (PSNR=29.26 dB); (e) LINC (PSNR=30.76 dB) (f) This work (GSM) (PSNR=32.09 dB).}
\label{fig:intexp}
\end{figure*}

\begin{figure*}
\subfloat[]{\includegraphics[width=.32\textwidth]{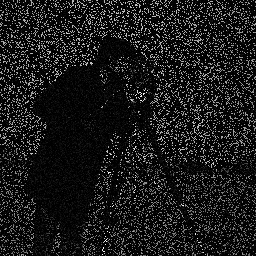}%
\label{(a)}}~
\subfloat[]{\includegraphics[width=.32\textwidth]{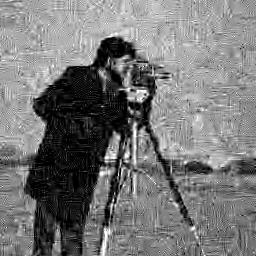}%
\label{(b)}}~
\subfloat[]{\includegraphics[width=.32\textwidth]{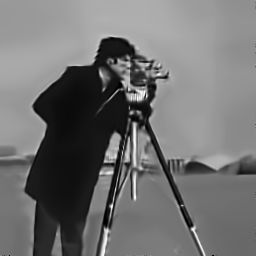}%
\label{(c)}}~
\caption{Example of restoring images with noisy and missing pixels. (a) Degraded image with 20\% of pixels available contaminated by Gaussian noise with $\sigma=30$ (b) EPLL (PSNR=21.02) (b) Proposed Method (PSNR=23.31).}
\label{fig:intnoisy}
\end{figure*}

An example of image interpolation with irregular missing data is shown in Fig. (\ref{fig:intexp}).
An example of recovering simultaneously missing and noisy pixels in high degraded regime is shown in Fig. (\ref{fig:intnoisy}).

\section{Conclusion}
In this paper, we proposed a CRF  to directly model the posterior probability distribution of the entire image to be estimated, given the observed image. Each potential function is considered as the product of a likelihood and a patch prior. We consider two priors in our frameworks: multivariate Gaussian and GSM. The experimental results show that the proposed method is competitive with the state-of-the-art denoising techniques, and outperforms the previous state-of-the-art in the case of image inpainting.

\appendices
\section{Proof of lemma 1}
Considering the fact that $|v_i\mathbf{\Sigma}_u|=v_i^n|\mathbf{\Sigma}_u|$, the optimization problem in \eqref{eq:zprop0} can be turned into the minimization problem of the form
\begin{eqnarray}
\hat{v}_i=\argmin_{v_i}\{ \frac{n}{2} \log v_i +{\frac{1}{2 v_i}{(\mathbf{z}_{i}-\sqrt{{v}_i} \boldsymbol{\mu}_{i})}^T \mathbf{\Sigma}_i^{-1}(\mathbf{z}_{i}-\sqrt{{v}_i} \boldsymbol{\mu}_{i})} \nonumber \\ +{{(1-\alpha)}\log v_i +\beta v_i}\}.
\label{eq:zop}
\end{eqnarray}
By renaming $d_i=\mathbf{z}_{i}^T \mathbf{\Sigma}_i^{-1}\mathbf{z}_{i}$ and $c_i=\mathbf{z}_{i}^T \mathbf{\Sigma}_i^{-1}\boldsymbol{\mu}_{i}$, and a simple rearranging, we have
\begin{equation}
\label{eq:zopsimp0}
{v}_i^{(new)}=\argmin_{v_i}{ \beta v_i+{{(1-\alpha+ \frac{n}{2})}\log v_i}+\frac{d_i}{2v_i}}-\frac{c_i}{\sqrt{v_i}}.
\end{equation}
Setting the derivate of the objective function in \eqref{eq:zopsimp0} to zero leads to
\begin{equation}
\beta + (1-\alpha+\frac{n}{2})\frac{1}{{v}_i}-\frac{d_i}{2 v_i^{2}} + \frac{c_i\, v_i^{-\frac{3}{2}}}{2} =0;
\end{equation}
since $v_i > 0$, by multiplying to $v_i^2$, this is equivalent to
\begin{equation}
\beta {v}_i^2+(1-\alpha+\frac{n}{2}){v}_i-\frac{d_i}{2}+\frac{c_i}{2} \sqrt{v_i}=0,
\end{equation}
which can be solved in closed form via a change of variables $v_i=w_i^2$,  yielding a quartic equation of the form
\begin{equation}
\beta {w}_i^4+(1-\alpha+\frac{n}{2}){w}_i^2+\frac{c_i}{2} w_i-\frac{d_i}{2}=0.
\end{equation}
Finally, the solution can be obtained by $v_i=\sqrt{w_i}$.

\section{Proof of lemma 2}
\begin{proof}
In this proof, for the simplicity of notations, we replace $\boldsymbol{\mu}_{\mathbf{u}_i}$ by $\boldsymbol{\mu}_{i}$. The covariance matrix of $\mathbf{z}_i$, denoted $\mathbf{C}_i$, is given by
\begin{equation}
\label{eq:covrelation}
\mathbf{C}_i=\mathbb{E}(v_i)\mathbb{E}(\mathbf{u}_i {\mathbf{u}_i}^T)-{\mathbb{E}^2(\sqrt{v_i})}\boldsymbol{\mu}_{i} \boldsymbol{\mu}_{i}^T
\end{equation}
Since  $v_i$ follows a Gamma distribution, then $\mathbb{E}(v_i) = \alpha/\beta$. On the other hand, it is known that if $v_i \sim Gamma{(\alpha,\beta)}$, then $\sqrt{v_i}$ follows a Nakagami distribution with expectation $E(\sqrt{v_i})= \sqrt{\beta} \, \Gamma(\alpha+\tfrac{1}{2})/\Gamma(\alpha)$.
Plugging these quantities in \eqref{eq:covrelation} leads to
\begin{equation}
\label{eq:covpara}
\mathbf{C}_i=\frac{\alpha}{\beta}\, \mathbb{E}(\mathbf{u}_i {\mathbf{u}_i}^T)-\Bigl( \frac{\Gamma(\alpha+\tfrac{1}{2})}{\Gamma(\alpha)}\sqrt{\beta}\Bigr)^2\boldsymbol{\mu}_{i} \boldsymbol{\mu}_{i}^T.
\end{equation} 
If 
\begin{equation}
\frac{\alpha}{\beta}= \Bigl(\frac{\Gamma(\alpha+\tfrac{1}{2})}{\Gamma(\alpha)}\sqrt{\beta}\Bigr)^2,\label{eq:cond_lemma3}
\end{equation}
then $\mathbf{C}_i = \tfrac{\alpha}{\beta} \bigl(  \mathbb{E}(\mathbf{u}_i {\mathbf{u}_i}^T) -  \boldsymbol{\mu}_{i} \boldsymbol{\mu}_{i}^T\bigr) = \tfrac{\alpha}{\beta} \mathbf{\Sigma}_i$. Finally, the condition in \eqref{eq:cond_lemma3} is equivalent to $\beta = \sqrt{\alpha}\; \Gamma(\alpha)/\Gamma(\alpha + \tfrac{1}{2})$.
\end{proof}

\bibliographystyle{IEEEtranS}
\bibliography{ref}

\end{document}